\documentclass{article}
\usepackage{spconf,graphicx}

\usepackage{amsfonts}       
\usepackage{nicefrac}       
\usepackage{microtype}      
\usepackage{url}

\usepackage[nosumlimits]{amsmath}
\usepackage{amssymb}
\usepackage{bm}
\usepackage{algorithm}
\usepackage{algpseudocode}

\usepackage{array}

\newcommand{\figref}[1]{{Fig.~\ref{#1}}}
\newcommand{\tabref}[1]{{Tab.~\ref{#1}}}
\newcommand{\secref}[1]{{Sec.~\ref{#1}}}
\newcommand{\asecref}[1]{{Appendix~\ref{#1}}}

\newcommand{\inputfig}[5][]{
\begin{figure#1}[tb]
\centering
\includegraphics[width=#2\linewidth]{#3}
\caption{#4}
\label{#5}
\end{figure#1}
}
\newcommand{\etc}{{\textit{etc}}}
\newcommand{\eg}{{\textit{e.g.}}}
\newcommand{\ie}{{\textit{i.e.}}}
\newcommand{\etal}{{\textit{et al.}}}

\newcommand{\statmu}[2][]{#1{\bm{\mu}}^\text{#2}}
\newcommand{\statsigma}[2][]{#1{\bm{\Sigma}}^\text{#2}}

\title{Covariance-aware Feature Alignment with Pre-computed Source Statistics for Test-time Adaptation to Multiple Image Corruptions}
\name{Kazuki Adachi$^\star$ \qquad Shin'ya Yamaguchi$^{\star\dagger}$ \qquad Atsutoshi Kumagai$^\star$}
\address{$^\star$NTT Computer and Data Science Laboratories \\
$^\dagger$Kyoto University \\
\texttt{\small \{kazuki.adachi,shinya.yamaguchi,atsutoshi.kumagai\}@ntt.com}
}

\begin{document}
\ninept
\maketitle
\begin{abstract}
Real-world image recognition systems often face corrupted input images, which cause distribution shifts and degrade the performance of models.
These systems often use a single prediction model in a central server and process images sent from various environments, such as cameras distributed in cities or cars.
Such single models face images corrupted in heterogeneous ways in test time.
Thus, they require to instantly adapt to the multiple corruptions during testing rather than being re-trained at a high cost.
Test-time adaptation (TTA), which aims to adapt models without accessing the training dataset, is one of the settings that can address this problem.
Existing TTA methods indeed work well on a single corruption.
However, the adaptation ability is limited when multiple types of corruption occur, which is more realistic.
We hypothesize this is because the distribution shift is more complicated, and the adaptation becomes more difficult in case of multiple corruptions.
In fact, we experimentally found that a larger distribution gap remains after TTA.
To address the distribution gap during testing, we propose a novel TTA method named \emph{Covariance-Aware Feature alignment (CAFe)}.
We empirically show that CAFe outperforms prior TTA methods on image corruptions, including multiple types of corruptions.

\end{abstract}

\section{Introduction}\label{sec:introduction}
\begin{figure}[tb]
\centering
\begin{tabular}{cc}
\includegraphics[width=0.4\linewidth]{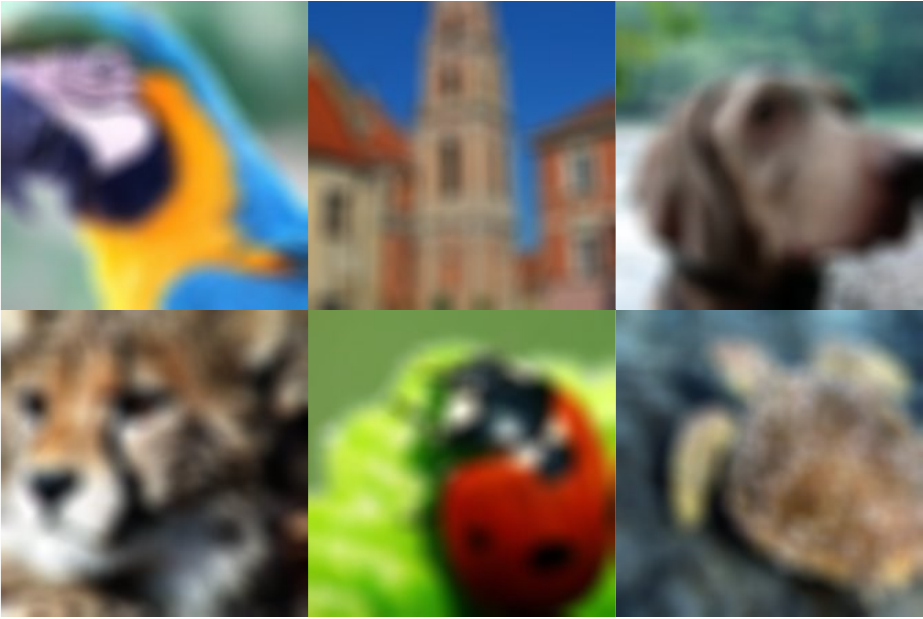}
&
\includegraphics[width=0.4\linewidth]{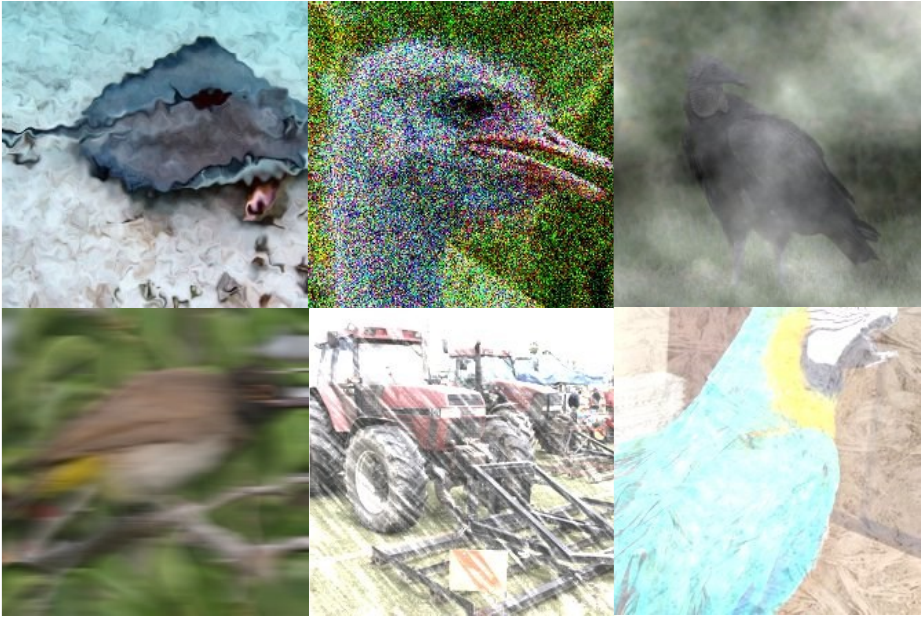} \\
{\small (a) single corruption (blur)} & {\small \shortstack{(b) multiple corruptions \\ (blur, noise, etc.)}}
\end{tabular}
\caption{Examples of (a) single and (b) multiple types of image corruptions.
(a) and (b) is composed of a single type of corruption and multiple types of corruptions, respectively.
We address to adapt to the target domain composed of multiple types of corruptions like (b), while prior TTA works mainly consider a single type of corruptions like (a).}
\label{fig:corruption_example}
\end{figure}

In real-world image recognition, the test distribution often changes from the training one because of corruptions such as noise or quality degradation caused by weather, location, \etc.
Such corruptions make models degrade their performance \cite{imagenet-c,recht2019imagenet}.
Vision systems such as surveillance cameras~\cite{sreenu2019intelligent} often use a single model in a central server and process images sent from various environments, such as cameras distributed in cities or cars.
In such situations, the model faces images corrupted in various ways dynamically depending on the environments surrounding the cameras.
For example, as illustrated in \figref{fig:corruption_example}, some images are foggy because of weather, others are block-noised because of narrowband, yet others are blurry because of a moving camera, and so on.
Since the corruptions change dynamically depending on environment, the model should adapt to the corruptions instantly rather than being re-trained on a labeled dataset like fine-tuning at a high cost.

\inputfig{0.8}{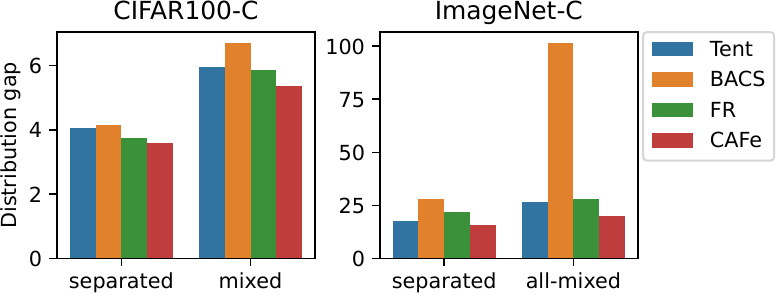}{Fr\'{e}chet distances of the features between the source and target datasets after TTA.
CAFe most reduces the distribution gap compared to other TTA methods on both separated (single corruption) and mixed (multiple corruptions) target datasets.
}{fig:feature_distance}

To adapt models to the test environment (target domain) in such a situation, recent studies have addressed a setting called \emph{test-time adaptation (TTA)}, which aims to adapt a model pre-trained in the training environment (source domain) to the target domain with an unlabeled target dataset in test time~\cite{schneider2020improving,li2017revisiting,Benz_2021_WACV,Wang2021,zhou2021bayesian,iwasawa2021test}.
This setting helps adapt a model trained on the source domain to the target domain dynamically in test time.
Prior TTA methods mainly focus on adapting batch normalization (BN) layers or refining output predictions.
Methods that adapt BN layers~\cite{schneider2020improving,li2017revisiting,Benz_2021_WACV} update the dimension-wise statistics (mean and variance) of the BN layers~\cite{batch-norm}.
On the other hand, methods that refine the predictions of the models \cite{Wang2021,zhou2021bayesian,iwasawa2021test} mainly focus on refining the confidence of the predictions such as entropy on the target domain~\cite{Wang2021,zhou2021bayesian} or modifying the weights of the final fully-connected layer~\cite{iwasawa2021test}.
They indeed work well on a single type of image corruption.
However, they degrade performance when images  corrupted in various ways come simultaneously, like the above case, which is a more realistic setting.
For example, Schneider \etal~\cite{schneider2020improving} report that adapting BN layers works well on a single type of image corruption but fails to adapt when corruptions are mixed.
Furthermore, we experimentally found that the performance of prior TTA methods other than those that adapt BN layers is also limited on multiple types of corruptions (see \secref{sec:experiment}).
We hypothesize that this is because the distribution shift is more complicated, and the adaptation becomes more difficult in case of multiple corruptions.
In fact, we experimentally observed that a larger distribution gap remains after TTA in case of multiple corruptions (See \figref{fig:feature_distance} and \secref{ssec:feature_alignment_effect}).
The distribution gap is crucial because it is included in the upper bound of the error on the target domain according to the theory of unsupervised domain adaptation~\cite{ben2010theory}.
However, the prior works optimize metrics that are hardly related to the distribution gap, such as entropy, or incorporate BN statistics that do not sufficiently represent the source distribution.

To address the distribution gap, we propose a novel TTA method called \emph{Covariance-Aware Feature alignment (CAFe)}.
CAFe aligns the feature distributions but does not access the source dataset.
To align the distributions precisely without accessing the source data, CAFe aligns the correlations between the dimensions of the feature vectors in addition to the mean and variance for each dimension.
To do this, we incorporate auxiliary statistics of the source dataset pre-computed before TTA.
Concretely, we pre-compute the mean and covariance of the features at training time in the source domain and then adapt the model to match the statistics on the target dataset during TTA.
This does not adversely affect the practicality of TTA because CAFe does not require accessing the source dataset itself during TTA.
One key challenge in CAFe is that the alignment fails when we naively compute the distribution gap with the source statistics.
This is because estimating the target distribution from a target mini-batch is difficult since the batch size is usually smaller than the number of feature dimensions (\eg, 2,048).
In such a case, the covariance matrix computed from the target mini-batch degenerates, which makes the adaptation unstable.
To address this issue, we propose \emph{feature grouping} to perform effective and stable feature alignment.
Feature grouping divides the dimensions of the feature representations into groups based on spectral clustering~\cite{alpert1995spectral} to find important correlations between the dimensions.
We avoid the degeneration of the covariance matrix by aligning the distributions for each group without ignoring important correlations between the dimensions.
We empirically show that CAFe outperforms the prior TTA methods on various image corruptions.
Moreover, we found that CAFe is effective, especially when the distribution shift consists of multiple types of corruption.

\section{Related Work}\label{sec:related_work}
\subsection{Unsupervised Domain Adaptation}\label{ssec:unsupervised_domain_adaptation}
Unsupervised domain adaptation (UDA) \cite{csurka2017domain} is an actively studied setting for adapting to distribution shifts.
Most UDA methods focus on minimizing the gap in feature distributions between the source and target domains to learn domain-invariant feature representations.
For example, DANN~\cite{ganin2016domain} and CORAL~\cite{sun2016deep,sun2016return} minimize the $\mathcal{H}$-divergence and the gap of the covariance matrices of the feature distributions, respectively.
These approaches are based on the fact that a term for the distribution gap is included in the expression of the upper bound of the generalization error on the target domain \cite{ganin2016domain,ben2010theory,Nguyen2022}.
However, UDA methods cannot be applied to the TTA setting because UDA requires a labeled source dataset and an unlabeled target dataset at the same time,
while TTA does not allow access to the source dataset in test time.

\subsection{Source-free Domain Adaptation}
Source-free domain adaptation (SFDA) is similar to TTA in terms of adapting to the target domain without accessing the source dataset.
SHOT~\cite{liang2020we} performs pseudo-labeling on the target dataset and then fine-tunes the model with the pseudo-labels.
Model adaptation~\cite{Li_2020_CVPR} uses a conditional GAN to generate target-style samples and fine-tunes the model on the generated labeled samples.
USFDA~\cite{kundu2020universal} generates artificial negative samples to learn tighter class boundaries during training.
The difference between SFDA and TTA is that SFDA is usually done in an offline manner for pseudo-labeling or training additional models.
In other words, SFDA requires the whole target dataset to be stored and be used for multiple epochs, which is a burden on storage and is computationally expensive.
On the other hand, TTA requires no additional models and can adapt in only one epoch or even in a mini-batched online manner.
This is computationally efficient since each target mini-batch can be discarded once it has been used to update a model and make a prediction.

\subsection{Test-time Adaptation}
Test-time adaptation (TTA) fine-tunes the source-pretrained model on the unlabeled target dataset like SFDA.
However, TTA does not require additional models or running for multiple epochs, unlike SFDA.
Thus, TTA can adapt to the target domain in a computationally efficient way.
Tent \cite{Wang2021} minimizes the entropy of the model predictions by optimizing the affine parameters of the BN layers.
BACS \cite{zhou2021bayesian} uses maximum-a-posteriori estimation and a model ensemble in addition to entropy minimization.
T3A \cite{iwasawa2021test} adjusts the weights of the last fully-connected layer by using the target samples.
However, these methods focus on refining the model outputs and lack the feature distribution alignment, although it is important to adapt to the target domain, as mentioned in \secref{ssec:unsupervised_domain_adaptation}.
On the other hand, adapting the BN layers (BN-adapt) \cite{schneider2020improving,li2017revisiting,Benz_2021_WACV} matches the BN statistics to the target ones.
However, Schneider \etal~\cite{schneider2020improving} report that BN-adapt works well on a single image corruption but fails to adapt to multiple corruptions.
More recently, feature restoration (FR)~\cite{eastwood2022sourcefree}, which stores the source feature distribution as a form of a histogram and aligns the target distribution to it, has been proposed.
Although CAFe is similar to FR in incorporating the source statistics, FR considers the dimension-wise distributions and does not capture the feature correlations.
In contrast to the prior works, we focus on accurate feature alignment inspired by UDA.
For this purpose, we seek to match the correlations in addition to the dimension-wise distributions by incorporating pre-computed source statistics.

\section{Problem Setting}
We first train a source model $f_\theta: \mathcal{X} \to \mathcal{Y}$ on a source dataset $\mathcal{S}=\{ (\mathbf{x}_i^\text{s},y_i^\text{s}) \} \sim p_\text{s}$, where $\mathcal{X}$ and $\mathcal{Y}$ are the input and label space, $\mathbf{x}^\text{s}_i\in \mathcal{X}$ and $y^\text{s}_i \in \mathcal{Y}$ are a training sample and its label in the source dataset, and $p_\text{s}$ is the source distribution over $\mathcal{X}\times \mathcal{Y}$.
Here, we can use arbitrary training or regularization methods.
The goal of TTA is to find the parameter $\theta^*$ that can make accurate predictions on the target domain with the pre-trained $f_\theta$ and an unlabeled target dataset $\mathcal{T}=\{ \mathbf{x}_i^\text{t} \} \sim p_\text{t}$, where $\mathbf{x}_i^\text{t} \in \mathcal{X}$ is a target sample and $p_\text{t}$ is the target distribution over $\mathcal{X}$.
The label space $\mathcal{Y}$ is shared between the two domains.
Note that the model cannot access the source dataset $\mathcal{S}$ during TTA, but we assume that the statistics (mean and covariance) of the feature representations pre-computed on $\mathcal{S}$ is available.
This does not undermine the advantages of TTA because we can compute the source statistics just after the training on $\mathcal{S}$.

\section{Covariance-aware Feature Alignment}
\inputfig{1}{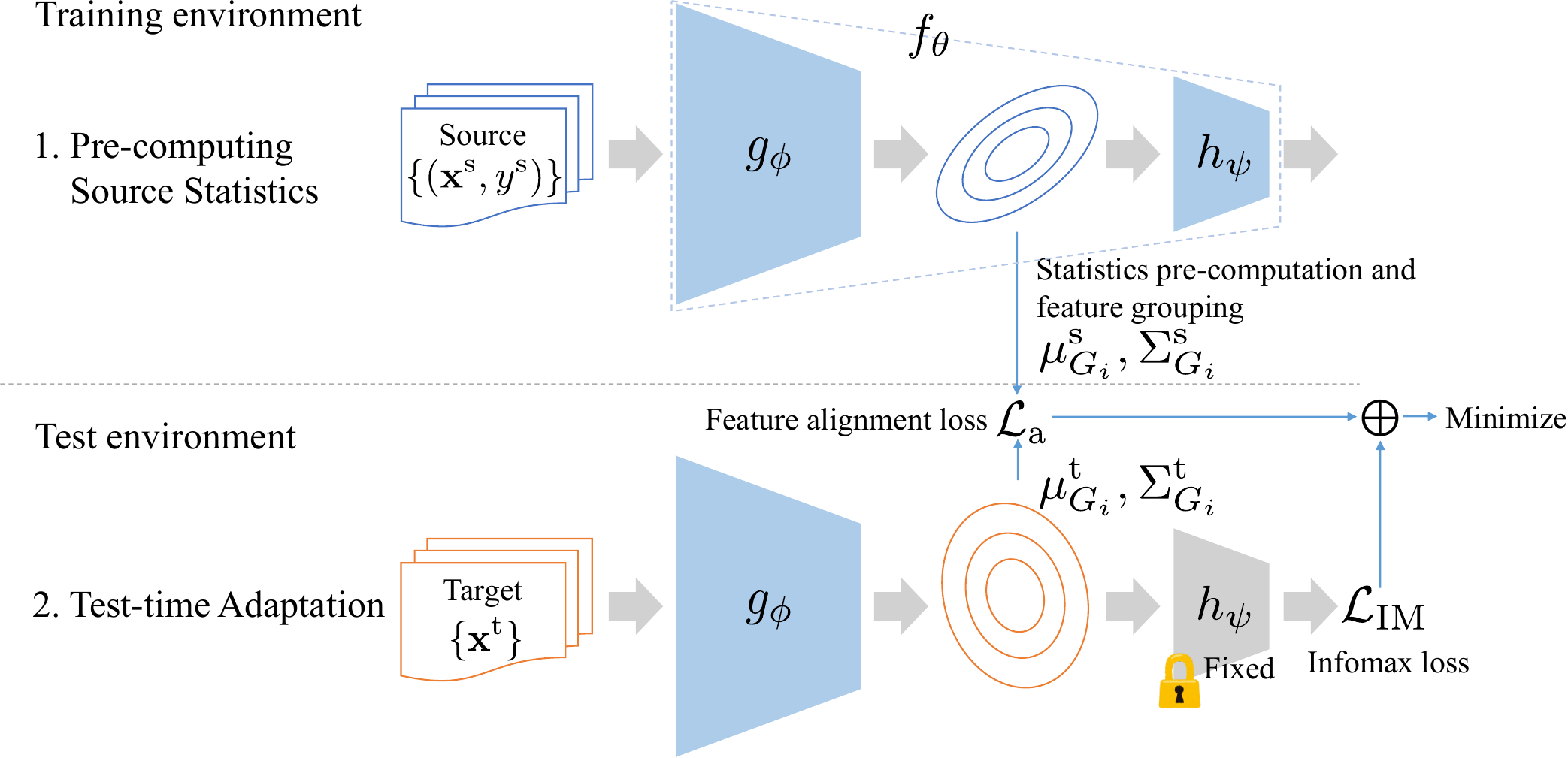}{Overview of CAFe.}{fig:cafe_overview}
\figref{fig:cafe_overview} illustrates an overview of CAFe.
CAFe is designed to make the target feature distribution close to the source one on the basis of the insights on UDA in \secref{ssec:unsupervised_domain_adaptation}.
We fine-tune $f_\theta$ to minimize the \emph{feature alignment loss} on an unlabeled target dataset $\mathcal{T}$ without accessing $\mathcal{S}$.
Using the pre-computed source statistics, feature alignment loss closes the distribution gap between the source and target domains, and the infomax loss encourages the feature representations to keep clusters that are hard to capture with the feature alignment loss (\secref{ssec:test-time_adaptation}).
However, if we compute the feature alignment loss naively, it will diverge because the covariance matrix of the target mini-batch degenerates.
Thus, we introduce \emph{feature grouping} to avoid the degenerate matrix and extract important correlations between the dimensions of the feature representations (\secref{ssec:feature_grouping}).

\subsection{Test-time Adaptation}\label{ssec:test-time_adaptation}
We split the model $f_\theta$ into a feature extractor $g_\phi : \mathcal{X} \to \mathbb{R}^d$ and classifier $h_\psi : \mathbb{R}^d \to \mathcal{Y}$, \ie, $f_\theta = h_\psi \circ g_\phi, \theta=[\psi, \phi]$.
After training $f_\theta$ on the source dataset $\mathcal{S}$, 
we calculate the mean vector $\statmu{s} \in \mathbb{R}^d$ and the covariance matrix $\statsigma{s} \in \mathbb{R}^{d\times d}$ of the feature representations $\mathbf{z}_i^\text{s} = g_\phi(\mathbf{x}_i^\text{s})$ on $\mathcal{S}$:
\begin{equation}
\statmu{s} \!=\! \frac{1}{|\mathcal{S}|} \sum_{i=1}^{|\mathcal{S}|} \mathbf{z}_i^\text{s}, ~~ \statsigma{s} \!=\! \frac{1}{|\mathcal{S}|} \sum_{i=1}^{|\mathcal{S}|} \left( \mathbf{z}_i^\text{s} - \statmu{s} \right) \! \left( {\mathbf{z}_i^\text{s}} -\statmu{s} \right)^{\!\top}. \label{eq:source_stats}
\end{equation}

Then, in the target domain, we perform TTA.
In this step, we cannot access the source dataset $\mathcal{S}$.
Here, we present the basic idea of CAFe and then modify it in \secref{ssec:feature_grouping}.
For each mini-batch of the target dataset $\mathcal{T}$, we compute the feature statistics $\statmu{t}$ and $\statsigma{t}$ analogously to Eq.~\eqref{eq:source_stats}.
Then, we compute the \emph{feature alignment loss} $\mathcal{L}_\text{a}$ to measure the discrepancy of the feature distributions by the KL-divergence of two multivariate Gaussians $\mathcal{N}^\text{t} =\mathcal{N}(\statmu{t} ,\statsigma{t} )$ and $\mathcal{N}^\text{s} =\mathcal{N}(\statmu{s} ,\statsigma{s})$.
This can be written in closed form~\cite{Duchi2007}:
\begin{eqnarray}
    D_\text{KL}(\mathcal{N}^\text{t}  \| \mathcal{N}^\text{s} ) = \frac{1}{2}\left[ \log \frac{\det (\statsigma{s}) }{\det (\statsigma{t})} - d + \text{tr}\left( {\statsigma{s} }^{-1} \statsigma{t}  \right) \right. \nonumber \\ 
     + \left. (\statmu{t} -\statmu{s} )^{\!\top} {\statsigma{s} }^{-1} (\statmu{t} -\statmu{s} ) \right]. \label{eq:kl-div}
\end{eqnarray}
We compute the feature alignment loss $\mathcal{L}_\text{a}$ by averaging the KL-divergence in both directions:
\begin{equation}
    \mathcal{L}_\text{a} = (1/2) \left[ D_\text{KL}(\mathcal{N}^\text{t} \| \mathcal{N}^\text{s}) + D_\text{KL}(\mathcal{N}^\text{s} \| \mathcal{N}^\text{t}) \right].
    \label{eq:feature_alignment_loss_naive}
\end{equation}

For measuring the discrepancy of the feature distributions above, we regard them as Gaussians ignoring classes.
Thus, to keep features clustered by classes, we use the infomax loss~\cite{NIPS1991_a8abb4bb}:
\begin{equation}
    \mathcal{L}_\text{IM} = -\frac{1}{B} \sum_{i=1}^B \sum_{j\in \mathcal{Y}}\hat{p}_{ij} \log \hat{p}_{ij} + \sum_{j\in \mathcal{Y}} \bar{p}_j\log \bar{p}_j, \label{eq:infomax_loss}
\end{equation}
where $B$ is the batch size, $\hat{p}_{ij}$ is the softmax probability of the prediction $f_\theta (\mathbf{x}_i^\text{t})$ for the $j$-th class, and $\bar{p}_j = (1/B)\sum_{i=1}^B \hat{p}_{ij}$.
The infomax loss is widely used in UDA and SFDA~\cite{liang2020we,vu2019advent,saito2019semi}.
It minimizes the entropy of the individual predictions to encourage data points to keep a certain distance from the decision boundaries. And it maximizes the entropy of the averaged probability to encourage the model to make diverse predictions.

For optimization, we fix the classifier $h_\psi$ and optimize the feature extractor $g_\phi$ to minimize $\mathcal{L}_\text{a}+\mathcal{L}_\text{IM}$.

\subsection{Feature Grouping}\label{ssec:feature_grouping}
The basic idea of CAFe is described in the previous section.
However, if we compute Eq.~\eqref{eq:kl-div} naively, the feature alignment loss $\mathcal{L}_\text{a}$ tends to diverge.
This is because the target covariance matrix $\statsigma{t}$ calculated over a mini-batch will degenerate, and $\det(\statsigma{t})$ becomes zero since the number of dimensions $d$ (\eg, 2,048 in ResNet-50~\cite{resnet}) is larger than the standard batch size.
A naive way to avoid the degenerate is adding a small constant to the diagonal elements of $\statsigma{t}$.
But this does not solve the degenerate problem intrinsically because the fact remains that the number of data in a mini-batch is insufficient to estimate the covariance.
Moreover, calculating the inverse of a large matrix (\eg, $2048\times 2048$), which is required by $\mathcal{L}_\text{a}$, is computationally expensive and generates significant numerical errors.

Hence, before computing $\mathcal{L}_\text{a}$, we perform \emph{feature grouping} to break down the dimensions of the feature representations $\{ 1,\ldots, d \}$ into groups smaller than the batch size and then compute $\mathcal{L}_\text{a}$ for each group.
Since the dimensions of the feature representations are not always independent, we want to split the dimensions into $k~(k<d)$ groups with strongly-correlated dimensions being put together as much as possible.
For this purpose, we use spectral clustering~\cite{alpert1995spectral}, which can split the nodes of a graph into clusters whose nodes are connected.
Here, we regard each dimension as a node of a graph and each correlation between the dimensions as a weight of an edge.
In other words, we consider a graph with a set of nodes $\mathcal{V}\!=\!\{1,\ldots, d\}$ and an adjacency matrix $A \in [0,1]^{d\times d}$ whose elements are the absolute values of the correlation coefficients of the feature dimensions in the source domain $A_{ij} \!=\! | \statsigma{s}_{ij} / \sqrt{\statsigma{s}_{ii} \statsigma{s}_{jj}} |.$
We perform spectral clustering on the graph to split the nodes (the dimensions) $\mathcal{V}$ into $k$ clusters (groups).
By doing so, the dimensions in the same group will have strong correlations, and the dimensions in different groups will have weak correlations.
We denote each group of dimensions by $G_1, \ldots, G_k$.
Then, we extract elements corresponding to each group $G_i$ from $\statmu{s}, \statsigma{s}, \statmu{t}, \statsigma{t}$, denoted by $\statmu{s}_{G_i}\!\in\! \mathbb{R}^{|G_i|}, \statsigma{s}_{G_i} \in \mathbb{R}^{|G_i|\times |G_i|}, \statmu{t}_{G_i}\in \mathbb{R}^{|G_i|}, \statsigma{t}_{G_i} \in \mathbb{R}^{|G_i|\times |G_i|}$, respectively.
The source statistics that we need to bring to the target domain are only $\statmu{s}$ and $\{\statsigma{s}_{G_i}\}_{i=1}^k$ instead of the whole $\statsigma{s}$.
The additional storage required in the target domain is negligible compared to the model size.

Instead of computing $\mathcal{L}_\text{a}$ for all dimensions at once, we compute it for each group $G_i$ with Eq.~\eqref{eq:feature_alignment_loss_naive} and then average them.
By feature grouping, we can make the number of dimensions for each group $|G_i|$ smaller than the batch size to prevent $\statsigma{t}_{G_i}$ from degenerating.

However, the source covariance $\statsigma{s}_{G_i}$ still may degenerate when only a small subspace of the feature space is used on the source domain.
To avoid this, we extract the principal components from $\statsigma{s}_{G_i}$ and compensate the degenerate directions before computing $\mathcal{L}_\text{a}$.
Like PCA, the principal components are eigenvectors of $\statsigma{s}_{G_i}$, denoted by $V_{G_i} \in \mathbb{R}^{|G_i|\times |G_i|}$, and the variances along the principal components are the corresponding eigenvalues $\lambda_{G_i}^1,\ldots,\lambda_{G_i}^{|G_i|}$.
To improve stability, we clip $\lambda_{G_i}^j$ with a small constant $\epsilon$, \ie, $\lambda_{G_i}^j \leftarrow \max ( \lambda_{G_i}^j, \epsilon)$.
Next, we map the source and target statistics to the space spanned by the source principal components $V_{G_i}$ to measure the KL-divergence with the modified source statistics:
\begin{eqnarray}
    \left( \statmu[\tilde]{t}_{G_i}, \statsigma[\tilde]{t}_{G_i} \right) &=&  \left( V_{G_i}(\statmu{t}_{G_i} - \statmu{s}_{G_i}), V_{G_i}^\top \statsigma{t}_{G_i} V_{G_i} \right), \label{eq:transformed_target_stats} \\
    \left( \statmu[\tilde]{s}_{G_i}, \statsigma[\tilde]{s}_{G_i} \right)  &=&  \left( V_{G_i}(\statmu{s}_{G_i} - \statmu{s}_{G_i}), V_{G_i}^\top \statsigma{s}_{G_i} V_{G_i} \right) \nonumber \\
     &=&  \left( \mathbf{0}, \text{diag}\left( \lambda_{G_i}^1,\ldots,\lambda_{G_i}^{|G_i|} \right) \right). \label{eq:transformed_source_cov}
\end{eqnarray}
Finally, we compute the KL-divergence in Eq.~\eqref{eq:kl-div} with these transformed statistics for each group.

\begin{table*}[t]
\centering
\begin{tabular}{l|ll|ll|lll}\hline
~ & \multicolumn{2}{c|}{CIFAR10-C} & \multicolumn{2}{c|}{CIFAR100-C} & \multicolumn{3}{c}{ImageNet-C} \\
Method & Separated & Mixed & Separated & Mixed & Separated & Severity-mixed & All-mixed \\ \hline
Source & $63.75$ & $63.46_{\pm 0.61}$ & $34.24$ & $34.16_{\pm 0.20}$  & $39.14$ & $39.43_{\pm 0.00}$ & $39.16_{\pm 0.01}$  \\
AdaBN~\cite{schneider2020improving} & $80.26_{\pm 0.30}$ & $67.62_{\pm 0.13}$ & $51.10_{\pm 0.25}$ & $38.52_{\pm 0.27}$ & $50.28_{\pm 0.02}$ & $48.00_{\pm 0.17}$ & $39.85_{\pm 0.18}$ \\
T3A~\cite{iwasawa2021test} & $66.02_{\pm 0.02}$ & $63.92_{\pm 0.42}$ & $36.05_{\pm 0.07}$ & $34.10_{\pm 0.49}$ & $39.05_{\pm 0.01}$ & $39.28_{\pm 0.03}$ & $37.46_{\pm 0.09}$ \\
Tent~\cite{Wang2021} & $80.86_{\pm 0.06}$ & $68.59_{\pm 0.30}$ & $52.09_{\pm 0.07}$ & $38.95_{\pm 0.65}$ & $58.97_{\pm 0.03}$ & $57.15_{\pm 0.05}$ & $44.44_{\pm 0.22}$  \\
BACS~\cite{zhou2021bayesian} & $\underline{81.51_{\pm 0.02}}$ & $68.69_{\pm 0.09}$ & ${\bf 53.00_{\pm 0.12}}$ & $39.65_{\pm 0.32}$ & $57.01_{\pm 0.19}$ & $55.05_{\pm 0.29}$ & $33.07_{\pm 1.38}$ \\
FR~\cite{eastwood2022sourcefree} & $80.71_{\pm 0.40}$ & $68.31_{\pm 0.64}$ & $51.50_{\pm 0.03}$ & $39.44_{\pm 0.32}$ & $53.54_{\pm 0.01}$ & $50.38_{\pm 0.20}$ & $40.52_{\pm 0.16}$ \\
Infomax~\cite{NIPS1991_a8abb4bb} & $81.40_{\pm 0.02}$ & $69.01_{\pm 0.50}$ & $52.48_{\pm 0.02}$ & $39.78_{\pm 0.36}$ & $60.20_{\pm 0.05}$ & $57.52_{\pm 0.23}$ & $46.52_{\pm 0.08}$ \\ \hline
CAFe (w/o infomax) & $81.11_{\pm 0.02}$ & $69.02_{\pm 0.62}$ & $51.83_{\pm 0.02}$ & $38.71_{\pm 0.11}$ & $57.35_{\pm 0.02}$ & $54.43_{\pm 0.14}$ & $43.83_{\pm 0.16}$ \\
CAFe (dimwise) & $81.40_{\pm 0.02}$ & $\underline{69.10_{\pm 0.38}}$ & $52.48_{\pm 0.02}$ & $\underline{39.83_{\pm 0.24}}$ & $\underline{60.29_{\pm 0.08}}$ & $\underline{58.60_{\pm 0.36}}$ & $\underline{47.19_{\pm 0.24}}$ \\
CAFe & ${\bf 81.66_{\pm 0.01}}$ & ${\bf 70.06_{\pm 0.25}}$ & $\underline{52.79_{\pm 0.02}}$ & ${\bf 40.01_{\pm 0.36}}$ & ${\bf 60.77_{\pm 0.09}}$ & ${\bf 59.04_{\pm 0.22}}$ & ${\bf 48.55_{\pm 0.26}}$ \\ \hline
\end{tabular}
\caption{Test accuracy [\%] on CIFAR10/100-C and ImageNet-C.
The accuracies on the separated sets are averaged over the corruption types.
For each column, the best and second-best accuracies are in {\bf bolded} and \underline{underlined}.}
\label{tab:accuracy}
\end{table*}

\section{Experiment}\label{sec:experiment}
We evaluated CAFe and other TTA methods on various types of image corruptions.
We found that CAFe produces better results and especially outperforms the other methods when the target domain consists of multiple types of corruption.

\subsection{Dataset}\label{ssec:dataset}
{\bf CIFAR}: We used CIFAR10/100 and CIFAR10/100-C~\cite{imagenet-c} as the source and target datasets.
CIFAR10/100-C are datasets that are shifted from CIFAR10/100 by applying 15 types of corruption at five severity levels to the test set of CIFAR10/100.
We used the images corrupted at the highest severity level.
First, we applied TTA to each type of corruption separately as in the previous works~\cite{Wang2021,zhou2021bayesian} (separated sets).
Then, we evaluated the TTA methods on a more severely shifted target dataset in which all 15 types of corruption are mixed (mixed set).
\\
{\bf ImageNet}: We used ImageNet \cite{lsvrc2012} as the source dataset and ImageNet-C \cite{imagenet-c} as the target dataset.
ImageNet-C is the ImageNet version of CIFAR10/100-C, wherein the same types of corruption are applied to the validation set of ImageNet.
Unlike the CIFAR experiment, we used images at all five severity levels.
First, we evaluated the TTA methods on each corruption and severity level separately as in the previous studies~\cite{Wang2021,zhou2021bayesian} (separated).
Second, we mixed the severity levels and evaluated the TTA methods on each corruption type (severity-mixed) to make the distribution shift harder.
Third, we mixed all the corruption types and severity levels (all-mixed).
The severity-mixed and all-mixed sets are more realistic distribution shifts because unknown and non-uniform corruptions can occur in the real world.

\subsection{Experimental Settings}\label{ssec:exp_setting}
We used ResNet-26 and ResNet-50~\cite{resnet} for CIFAR and ImageNet experiments.
We used the outputs of the last global-average pooling layer as the feature representations.
\\
{\bf CAFe}: We set the number of groups $k$ to 128.
For the feature grouping, we used {\tt sklearn.cluster.SpectralClustering} in Scikit-learn \cite{scikit-learn} with the default hyperparameters.
We optimized the feature extractor with momentum SGD for one epoch, a standard setting in TTA.
We set the learning rate to 0.001 and the momentum to 0.8.
We found that a larger batch size is better and set it to 256 since the target statistics can be estimated precisely.
{\bf Baselines}: We compare CAFe with these previous TTA methods and variants of CAFe.
We compare CAFe with existing TTA methods: AdaBN~\cite{schneider2020improving}, Tent~\cite{Wang2021}, BACS~\cite{zhou2021bayesian}, T3A~\cite{iwasawa2021test}, and FR~\cite{eastwood2022sourcefree}.
For ablation, we also compared with variants of CAFe: Infomax~\cite{NIPS1991_a8abb4bb}, CAFe (w/o infomax), and CAFe (dimwise).
Infomax and CAFe (w/o infomax) minimizes only $\mathcal{L}_\text{IM}$ and $\mathcal{L}_\text{a}$, respectively.
CAFe (dimwise) ignores the correlation between feature dimensions in computing $\mathcal{L}_\text{a}$.

We ran each TTA three times with different random seeds and reported the mean and standard deviations.

\subsection{Feature Distribution Gap}\label{ssec:feature_alignment_effect}
First, to check whether the multiple corruptions cause a larger distribution gap than the single ones, we measured how close the distribution gap between the source and target datasets became by TTA.
After TTA, we calculated the feature mean $\statmu{t}$ and covariance $\statsigma{t}$ for each target dataset and evaluated the distribution gap between the two domains as the Fr\'{e}chet distance between $\mathcal{N}(\statmu{s},\statsigma{s})$ and $\mathcal{N}(\statmu{t}, \statsigma{t})$.
\figref{fig:feature_distance} shows the distribution gap.
In the multiple corruptions case, a larger distribution gap remains after TTA.
But CAFe effectively closed the distribution gap.

\subsection{Test Accuracy}\label{ssec:cifar_result}
The left and center columns of \tabref{tab:accuracy} show the test accuracy on CIFAR10/100-C.
CAFe outperformed the baselines.
Even when the correlations between the features were ignored in $\mathcal{L}_\text{a}$, CAFe (dimwise) still produced competitive results.
CAFe had higher accuracy, especially on the mixed target datasets.
This indicates that CAFe enables the models to adapt to more realistic distribution shifts that various types of corruptions can occur.
On the other hand, CAFe (w/o infomax) had poorer accuracy compared with the other CAFe variants.
This suggests that regarding the feature distributions as multivariate Gaussians in $\mathcal{L}_\text{a}$ hardly captures clusters of features and that the infomax loss works complementarily.

The right columns of \tabref{tab:accuracy} show the test accuracy on ImageNet-C.
CAFe outperformed the baselines in all cases.
Especially in the all-mixed case, CAFe had significantly higher accuracy.
This implies that aligning the feature distributions becomes more important when the distribution shift is complicated.

\section{Conclusion}\label{sec:discussion}
We proposed a novel TTA method called Covariance-Aware Feature alignment (CAFe), 
which aligns the target feature distribution to the source distribution.
We also proposed feature grouping to extract important correlations between the feature dimensions by spectral clustering and improve computing stability.
Experimental results show that CAFe effectively closes the distribution gap in the feature space and improves the test accuracy on various types of image corruption.

\vfill\pagebreak

\bibliographystyle{IEEEbib}
\bibliography{0_main}

\clearpage

\appendix

\section{Implementation Details}\label{asec:implementation_details}
We implemented CAFe and the other methods with PyTorch \cite{pytorch} and PyTorch-Ignite \cite{pytorch-ignite}.

\subsection{Source Pre-training}\label{assec:source_pretraining}
ResNet-26 and ResNet-50 that we used are included in timm \cite{rw2019timm} and Torchvision~\cite{torchvision2016} libraries.

For CIFAR experiments, we trained ResNet-26 with momentum SGD from scratch.
We set the batch size to $128$, initial learning rate to $0.1$, momentum to $0.9$, and weight decay to $5\times 10^{-4}$.
We decayed the learning rate to zero with cosine annealing \cite{cosine_annealing} over the first $200$ epochs.
Then, from epoch $200$, we fixed the learning rate to $0.01$ and collected SWAG \cite{maddox2019simple} for $100$ epochs.
For SWAG, the parameters are collected four times per epoch.
We used the means of the collected parameters as the source model.

For ImageNet experiments, we downloaded the ImageNet-pretrained ResNet-50 via Torchvision.
For BACS \cite{zhou2021bayesian}, we further trained the ImageNet-pretrained model on ImageNet to collect the SWAG posterior.
We ran vanilla SGD for optimization with $\text{learning\_rate}=10^{-4}$ and $\text{batch\_size}=128$ for 10 epochs and collected the parameters four times per epoch.

\subsection{Adaptation}\label{assec:adaptation_details}
We describe the details of CAFe and other TTA methods.
For each condition, we ran adaptation three times with different random seeds and averaged the accuracies. \\
{\bf AdaBN}~\cite{schneider2020improving}: We ran the test run with keeping updating the BN statistics.
We set the batch size to 32.
\\
{\bf BACS}~\cite{zhou2021bayesian}: We set the hyperparameters same as the original paper~\cite{zhou2021bayesian}.
We used momentum SGD for optimization.
For CIFAR experiments, we set the batch size to 128, learning rate to 0.001, momentum to 0.8, and the weight of the posterior term to $10^{-4}$.
For ImageNet experiments, we set the batch size to 64, learning rate to $10^{-4}$, momentum to 0.9, and the weight of the posterior term to $3\times 10^{-4}$.
\\
{\bf Tent}~\cite{Wang2021}: We set the hyperparameters as the original paper~\cite{Wang2021}.
We used momentum SGD for optimization.
For CIFAR experiments, we set the batch size to 128, learning rate to 0.001, and momentum to 0.8.
For ImageNet experiments, we set the batch size to 64, learning rate to $2.5\times 10^{-4}$, and momentum to 0.8.
\\
{\bf T3A}~\cite{iwasawa2021test}: We set the batch size to 32 as the original paper~\cite{iwasawa2021test}.
For the number of feature vectors to pool $M$, we searched from $\{1, 5, 20, 50, 100, \infty\}$ and adopted $M=20$ that consistently performs the best for all experiments.
\\
{\bf FR}~\cite{eastwood2022sourcefree}: We omit the bottom-up FR since it cannot be applied to TTA setting in our experiment.
As the original paper~\cite{eastwood2022sourcefree}, we set the number of bins to 8 and the soft binning temperature parameter to 0.01 for offline and 0.05 for the online setting.
For optimization, we used momentum SGD and set the learning rate to 0.001 and the momentum to 0.9, and the batch size to 256.
\\
{\bf CAFe (ours)}: The hyperparameters are described in \secref{ssec:exp_setting}.
For computing the KL-divergence in Eq.~\eqref{eq:kl-div}, we made some modifications for stability.
The term $\log \det( \statsigma{s}_{G_i} )/ \det( \statsigma{t}_{G_i} ) = \log \det( \statsigma{s}_{G_i} ) - \log \det( \statsigma{t}_{G_i} )$ may diverge if we compute it naively because the determinant of the covariance matrices often become near-zero.
Thus, we computed the log of the covariance matrix with Cholesky decomposition: 
\begin{equation}
    \log \det(\statsigma{t}_{G_i}) = 2\sum_{j=1}^{|G_i|} \log L_{jj},
\end{equation}
where $L$ is a lower triangular matrix obtained by Cholesky decomposition such that $LL^\top=\statsigma{t}_{G_i}$.
$\log \det(\statsigma{s}_{G_i})$ is computed as the following since  $\statsigma{s}_{G_i}$ becomes a diagonal matrix after the transformation in Eq.~\eqref{eq:transformed_source_cov}:
\begin{equation}
    \log \det(\statsigma{s}_{G_i}) = \sum_{j=1}^{|G_i|} \log \lambda_{G_i}^j.
\end{equation}

\subsection{Fr\'{e}chet Distance} \label{assec:frechet_distance}
In Sec. 5.5, we measured the distance between the source and target distributions by using the Fr\'{e}chet distance between the two Gaussians $\mathcal{N}(\statmu{t}, \statsigma{t})$ and $\mathcal{N}(\statmu{s}, \statsigma{s})$, which is calculated as~\cite{DOWSON1982450}:
\begin{equation}
\| \statmu{t} - \statmu{s} \|_2^2 + \text{tr} \left( \statsigma{t} + \statsigma{s} - 2 (\statsigma{t}\statsigma{s})^{1/2} \right).
\end{equation}
\section{Dataset Details}\label{asec:dataset_details}
{\bf CIFAR10/100} \cite{krizhevsky2009learning}: We downloaded CIFAR10/100 via Torchvision.
We could not find the license information for CIFAR10/100.

{\bf ImageNet} \cite{lsvrc2012}: We downloaded ImageNet from the official site (\url{https://www.image-net.org/}).
ImageNet is released under a license that allows it to be used for non-commercial research/educational purposes (see the official site).

{\bf CIFAR10/100-C and ImageNet-C} \cite{imagenet-c}: We downloaded CIFAR10/100-C and ImageNet-C from the following URLs, which are released by the authors:
\begin{itemize}
\item \url{https://zenodo.org/record/2535967}
\item \url{https://zenodo.org/record/3555552}
\item \url{https://zenodo.org/record/2235448}
\end{itemize}
They are released under the CC BY 4.0 License.

{\bf SVHN}~\cite{svhn}: We downloaded SVHN via Torchvision.
It can be used for non-commercial purposes only (see \url{http://ufldl.stanford.edu/housenumbers/}).

{\bf MNIST}~\cite{mnist}: We downloaded MNIST via Torchvision.
We could not find the license information for MNIST.
\renewcommand{\algorithmicrequire}{\textbf{Input:}}
\renewcommand{\algorithmicensure}{\textbf{Output:}}

\section{Algorithm of CAFe} \label{asec:cafe_algo}
Alg.~\ref{aalg:cafe_alg_source_stats} and Alg.~\ref{aalg:cafe_alg_tta} describe the procedure of CAFe.

\begin{algorithm}
\caption{Algorithm of CAFe (Pre-computing source statistics).}
\label{aalg:cafe_alg_source_stats}
\begin{algorithmic}
\State{{\#} These steps are done in the source environment.}
\Require Source-pretrained model $f_\theta=h_\psi \circ g_\phi$, source dataset $\{ \mathbf{x}_i^\text{s} \}$
\Ensure Feature group $G_1,\ldots, G_k$, source feature statistics for each group $\{ \statmu{s}_{G_i} \}_{i=1}^k, \{ \statsigma{s}_{G_i} \}_{i=1}^k$

\State{Extract feature $\mathbf{z}_i^\text{s} =  g_\phi(\mathbf{x}_i^\text{s})$ for each sample.}
\State{Compute the source feature statistics $\statmu{s}$ and $\statsigma{s}$ with Eq.~\eqref{eq:source_stats}.}
\State{Compute the adjacency matrix $A_{ij}=|\statsigma{s}_{ij}/\sqrt{\statsigma{s}_{ii}\statsigma{s}_{jj}}|$.}
\State{Split the feature dimensions $\mathcal{V}=\{ 1,\ldots, d \}$ into $k$ groups $G_1,\ldots, G_k$ by spectral clustering according to the adjacency matrix $A$.}
\State{Extract elements corresponding to each $G_i$ from $\statmu{s}$ and $\statsigma{s}$, denoted by $\statmu{s}_{G_i}$ and $\statsigma{s}_{G_i}$.}
\end{algorithmic}
\end{algorithm}

\begin{algorithm}
\caption{Algorithm of CAFe (TTA step).}
\label{aalg:cafe_alg_tta}
\begin{algorithmic}
\State{{\#} These steps are done in the target environment without access to the source dataset.}
\Require Source-pretrained model $f_\theta=h_\psi \circ g_\phi$, unlabeled target dataset $\{ \mathbf{x}_i^\text{t} \}$, \\
feature group $G_1,\ldots, G_k$,
source feature statistics for each group $\{ \statmu{s}_{G_i} \}_{i=1}^k, \{ \statsigma{s}_{G_i} \}_{i=1}^k$
\Ensure Target-adapted model $f_{\theta'}$

\State{Compute the eigenvalues $\lambda^1_{G_i},\ldots, \lambda^{|G_i|}_{G_i}$ and eigenvectors $V_{G_i}$ of the source covariance matrix $\statsigma{s}_{G_i}$ for each $G_i$.}
\State{Clip the eigenvalues.}
\State{Compute the transformed source statistics $\statmu[\tilde]{s}_{G_i}$ and $\statsigma[\tilde]{s}_{G_i}$ with Eq.~\eqref{eq:transformed_source_cov}.}
\For{each mini-batch $\{ \mathbf{x}_i^\text{t} \}_{i=1}^B$}
  \State{Extract feature $\mathbf{z}_i^\text{t} =  g_\phi(\mathbf{x}_i^\text{t})$ and make prediction $\hat{p}_i =\text{softmax}(h_\psi(\mathbf{z}_i^\text{t}))$ for each sample.}
  \State{Compute the batch statistics $\statmu{t}$ and $\statsigma{t}$ analogously to Eq.~\eqref{eq:source_stats}.}
  \State{Extract elements from $\statmu{t}$ and $\statsigma{t}$ corresponding to each $G_i$, denoted by $\statmu{t}_{G_i}$ and $\statsigma{t}_{G_i}$.}
  \State{Compute the transformed target statistics $\statmu[\tilde]{t}_{G_i}$ and $\statsigma[\tilde]{t}_{G_i}$ with  Eq.~\eqref{eq:transformed_target_stats}.}
  \State{Compute the feature alignment loss $\mathcal{L}_\text{a}$ with the transformed statistics $\statmu[\tilde]{t}_{G_i}, \statsigma[\tilde]{t}_{G_i}, \statmu[\tilde]{s}_{G_i}, \statsigma[\tilde]{s}_{G_i}$ by Eq.~\eqref{eq:feature_alignment_loss_naive}.}
  \State{Compute the infomax loss $\mathcal{L}_\text{IM}$ with Eq.~\eqref{eq:infomax_loss}.}
  \State{Update the feature extractor $g_\phi$ to minimize $\mathcal{L}_\text{a} + \mathcal{L}_\text{IM}$.}
\EndFor
\end{algorithmic}
\end{algorithm}
\section{Additional Experiment} \label{asec:additional_experiment}
\subsection{Detailed TTA Results} \label{assec:adaptation_result}
\inputfig[*]{1}{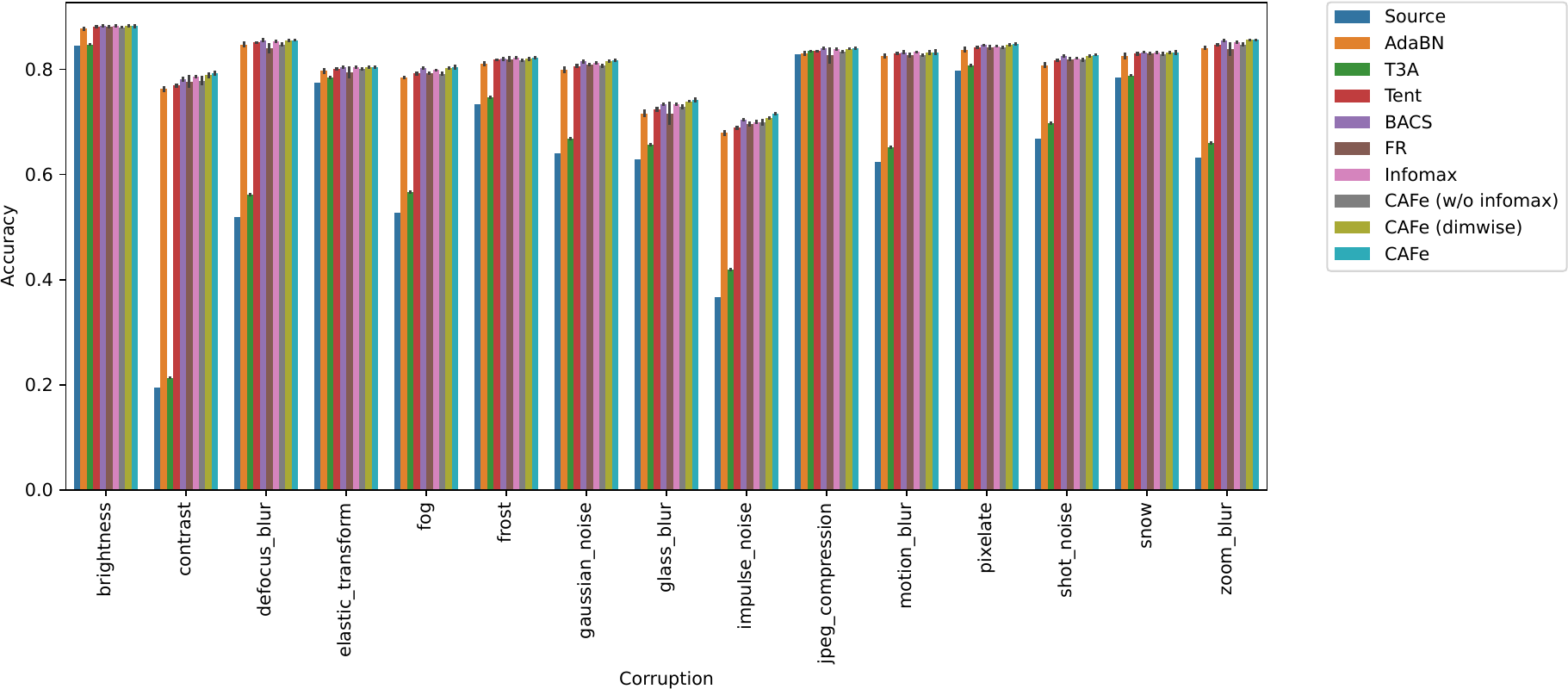}{Test accuracy for each corruption type on the separated sets of CIFAR10-C.}{afig:cifar10c_domain-wise_accuracy}

\inputfig[*]{1}{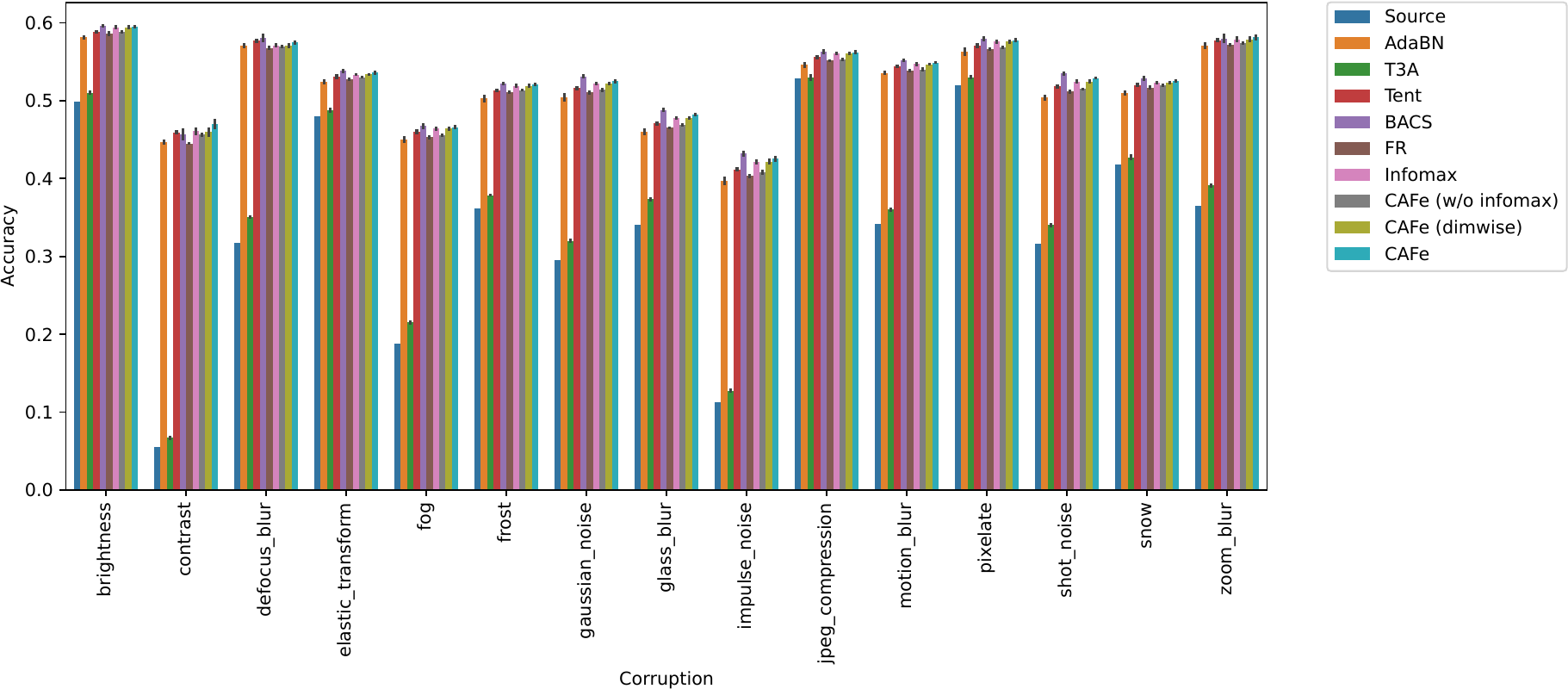}{Test accuracy for each corruption type on the separated sets of CIFAR100-C.}{afig:cifar100c_domain-wise_accuracy}

\inputfig[*]{1}{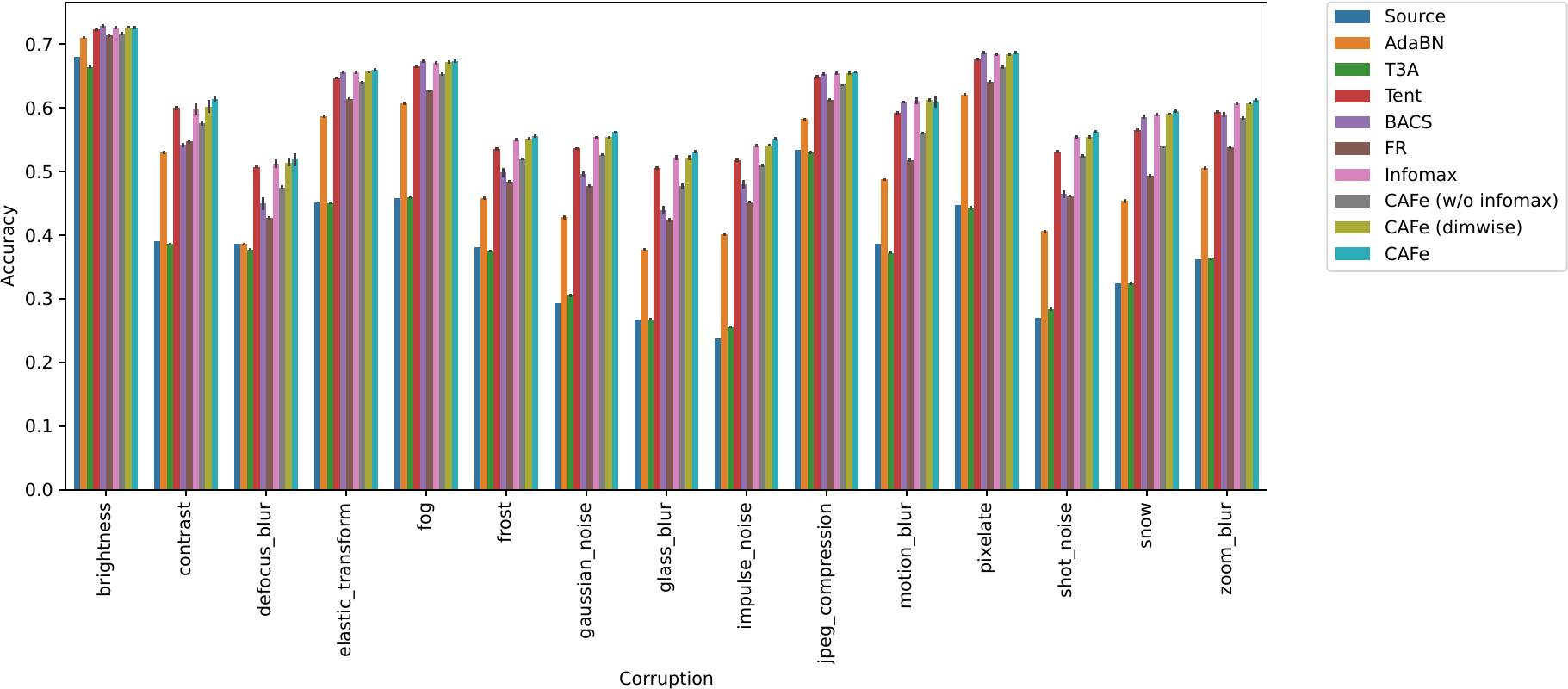}{Test accuracy for each corruption type on the severity-wise sets of ImageNet-C.
The accuracies are averaged over severity levels.}{afig:imagenet-c_severity-wise_accuracy}

\inputfig[*]{1}{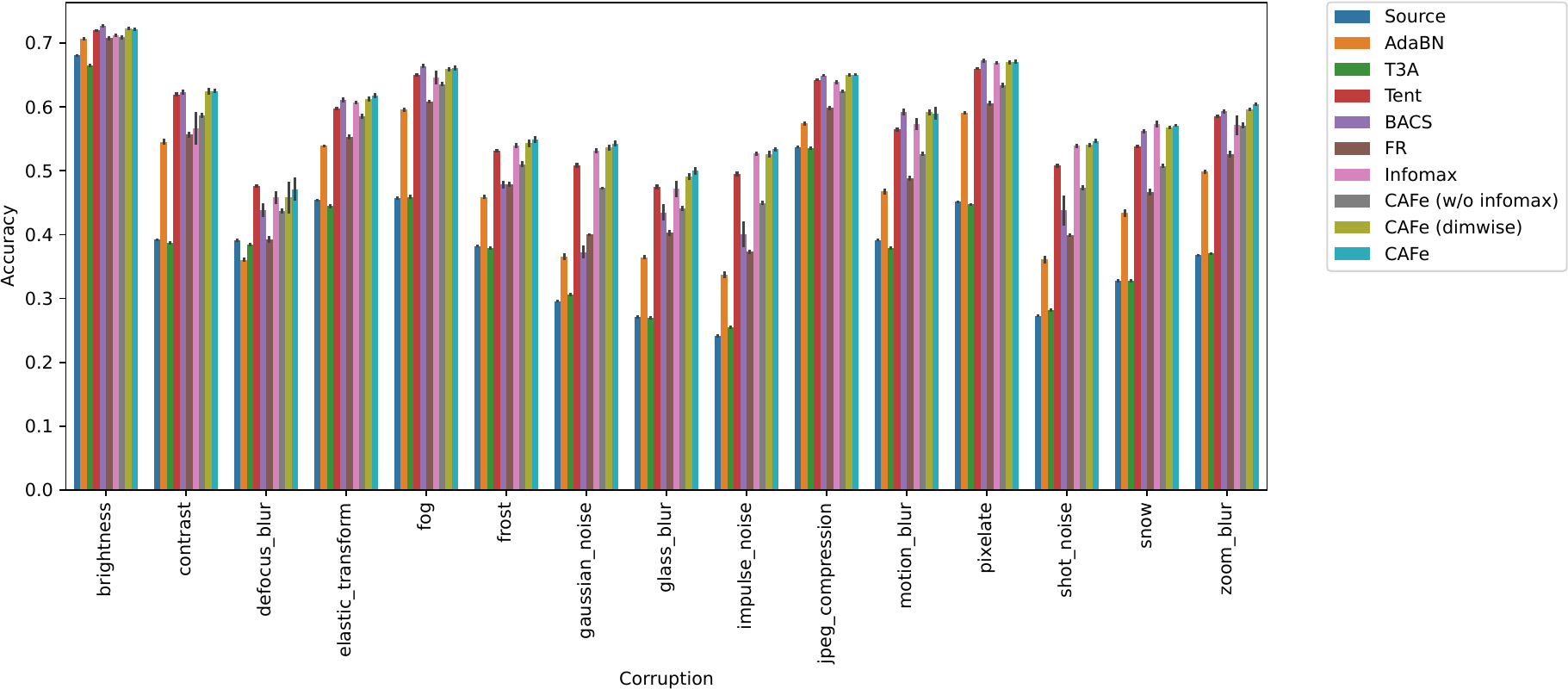}{Test accuracy for each corruption type on the severity-mixed sets of ImageNet-C.}{afig:imagenet-c_severity-mixed_accuracy}

\figref{afig:cifar10c_domain-wise_accuracy} -- \figref{afig:imagenet-c_severity-mixed_accuracy} show the detailed test accuracies on CIFAR-10/100-C and ImageNet-C reported in \tabref{tab:accuracy} for each corruption type.

\subsection{Effect of the Feature Grouping}\label{assec:feature_grouping_result}
\begin{table}[tb]
\centering
{\small
\begin{tabular}{l|ll}\hline
$k$ & \shortstack{CIFAR100-C \\ (Mixed)} & \shortstack{ImageNet-C \\ (All-mixed)} \\ \hline
8 & N/A & N/A \\ 
16 & N/A & $0.11 \pm 0.02$ \\ 
32 & ${\bf 41.06 \pm 0.18}$ & $46.22 \pm 0.16$ \\ 
64 & $40.67 \pm 0.21$ & ${\bf 48.58 \pm 0.13}$ \\ 
128 & $40.01 \pm 0.36$ & $48.55 \pm 0.26$ \\ 
256 & $39.60 \pm 0.10$ & $48.21 \pm 0.31$ \\ 
512 & N/A & N/A \\ 
1024 & $39.70 \pm 0.09$ & N/A \\ \hline
\end{tabular}
}
\caption{Test accuracy of CAFe on CIFAR100-C (Mixed) and ImageNet-C (All-mixed) for each group size $k$.
The best accuracy on each dataset is {\bf bolded}.
N/A means that CAFe did not work because of the degenerate of the  covariance matrix.
}
\label{atab:feature_grouping}
\end{table}

We examined the effect of the number of groups $k$ in the feature grouping.
\tabref{atab:feature_grouping} shows the test accuracy of CAFe on CIFAR100-C (Mixed) and ImageNet-C (All-mixed) for each group size $k$.
$k$ affects the number of correlations between the feature dimensions to be considered.
Using smaller $k$, more correlations are considered because the size of each group $G_i$ increases.
In the case of CIFAR100-C (Mixed), $k=32$ gives in the best accuracy.
This is because the distributions are aligned accurately.
However, when $k\leq 16$, CAFe does not work because some of the feature groups are larger than the batch size.
For $G_i$ larger than the batch size, the KL-divergence in Eq.~\eqref{eq:kl-div} diverges since $\text{det}(\statsigma{t}_{G_i}) = 0$ for a mini-batch.
For ImageNet-C (All-mixed), $k=64$ gives the best accuracy.
In the $k\leq 32$ cases, the accuracies decrease.
This is because the covariance estimation becomes inaccurate as the number of dimensions of the feature groups increases.
On the other hand, in some cases of larger $k$, the spectral clustering produces a few large feature groups and many small ones.
Thus, the KL-divergences for the large groups diverge.
However, we can also see that the choice of $k$ does not largely affect accuracy in most cases when CAFe works.
Practically, we recommend choosing $k$ such that the feature group size is sufficiently smaller than the batch size $B$, i.e., $k> d/B$.
The relationship between the size of the feature groups and $k$ is visualized in Sec. C.2 in the supplementary materials.

\begin{figure*}[tb]
\centering
\begin{tabular}{cc}
\includegraphics[height=30em]{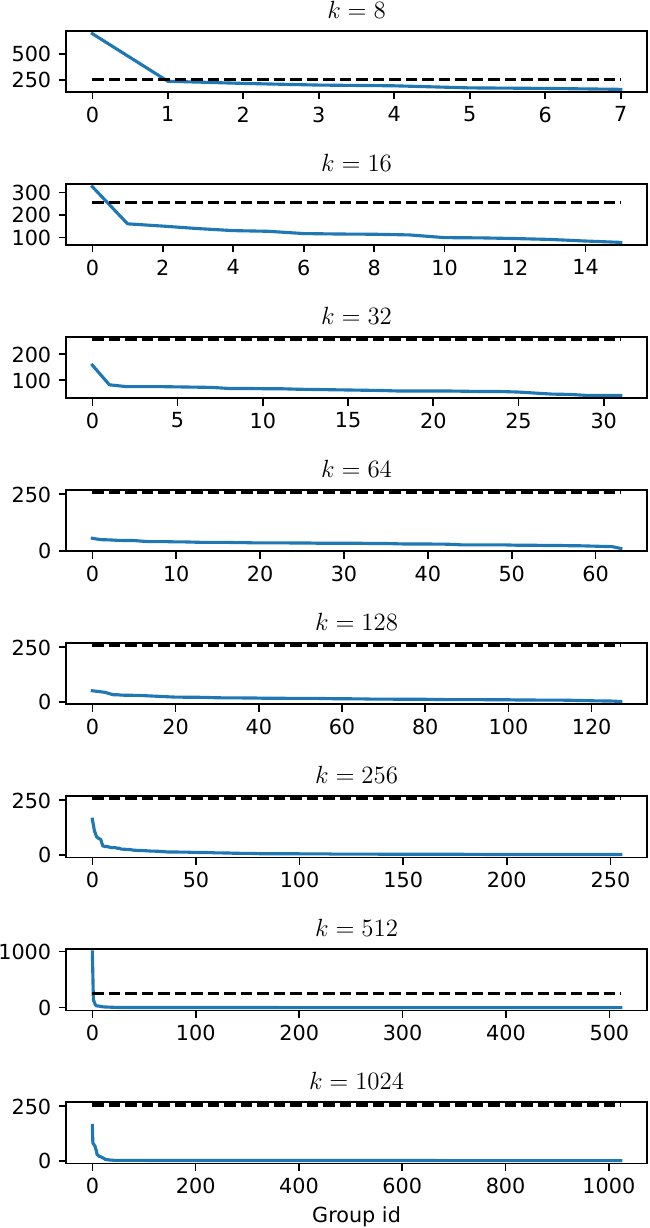}
&
\includegraphics[height=30em]{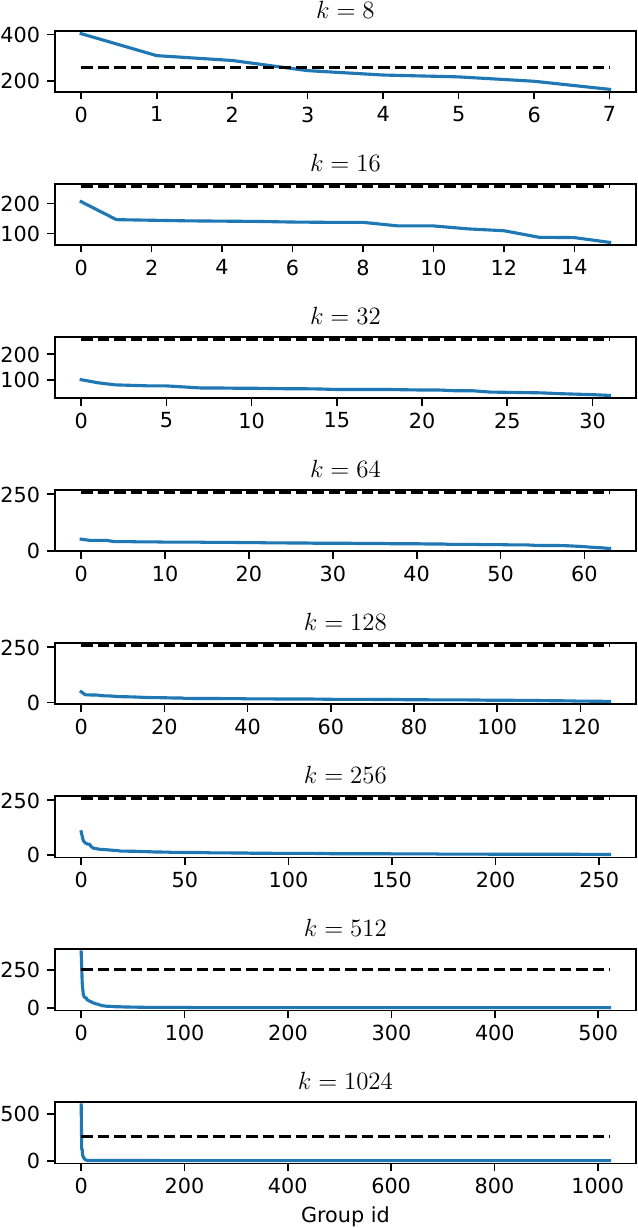} \\
(a) CIFAR100 & (b) ImageNet
\end{tabular}
\caption{Size of each group of feature dimensions for each $k$.
The group ids are sorted in descending order of group size.
The blue and dashed black lines indicate the group size and batch size ($=256$).}
\label{afig:cluster_size}
\end{figure*}

Further, we inspected the size of each group made by feature grouping.
\figref{afig:cluster_size} shows the size of each group for each $k$ on CIFAR100 and ImageNet.
When $k\in \{8,16,512\}$ on CIFAR100 and $k\in \{8,512,1024\}$ on ImageNet, groups that are larger than the batch size occur.
In these cases, the target covariance matrix $\statsigma{t}$ for the mini-batch degenerates and the feature alignment loss $\mathcal{L}_\text{a}$ diverges.
Such cases are correspond to N/A in \tabref{atab:feature_grouping}.

\subsection{Mini-batched Online TTA}\label{assec:online_tta}
\begin{table*}[tb]
\centering
\begin{tabular}{l|ll|ll} \hline
~ & \multicolumn{2}{c|}{CIFAR10-C} & \multicolumn{2}{c}{CIFAR100-C} \\
Method & Separated & Mixed & Separated & Mixed \\ \hline
Source & $63.75$ & $63.46 \pm 0.61$ & $34.24$ & $34.16 \pm 0.20$ \\
AdaBN~\cite{schneider2020improving} & ${\bf 80.04 \pm 0.03}$ & $67.94 \pm 0.16$ & ${\bf 50.72 \pm 0.01}$ & $38.20 \pm 0.09$ \\
T3A~\cite{iwasawa2021test} & $66.02 \pm 0.02$ & $63.92 \pm 0.42$ & $36.05 \pm 0.07$ & $34.10 \pm 0.49$ \\
Tent~\cite{Wang2021} & $79.31 \pm 0.05$ & $68.57 \pm 0.25$ & $50.04 \pm 0.07$ & $\underline{39.00 \pm 0.09}$ \\
BACS~\cite{zhou2021bayesian} & $\underline{79.67 \pm 0.05}$ & $68.28 \pm 0.47$ & $\underline{50.60 \pm 0.07}$ & $38.98 \pm 0.06$ \\
FR~\cite{eastwood2022sourcefree} & $77.87 \pm 0.11$ & $68.65 \pm 0.87$ & $47.93 \pm 0.05$ & $38.70 \pm 0.01$ \\
Infomax~\cite{NIPS1991_a8abb4bb} & $78.04 \pm 0.10$ & $68.80 \pm 0.62$ & $48.49 \pm 0.05$ & $38.62 \pm 0.37$ \\ \hline
CAFe (w/o infomax) & $77.90 \pm 0.11$ & $\underline{69.08 \pm 0.34}$ & $48.13 \pm 0.03$ & $38.59 \pm 0.37$ \\
CAFe (dimwise) & $78.04 \pm 0.10$ & $68.91 \pm 0.34$ & $48.50 \pm 0.06$ & $38.86 \pm 0.20$ \\
CAFe & $78.18 \pm 0.13$ & ${\bf 69.34 \pm 0.43}$ & $48.64 \pm 0.03$ & ${\bf 39.22 \pm 0.28}$ \\ \hline
\end{tabular}
\caption{Test accuracy [\%] on CIFAR10/100-C with mini-batched online TTA.}
\label{atab:online_cifar_accuracy}
\end{table*}

\begin{table*}[tb]
\centering
\begin{tabular}{l|lll}\hline
Method & Separated & Severity-mixed & All-mixed \\ \hline
Source & $39.14$ & $39.43 \pm 0.00$ & $39.16 \pm 0.01$ \\
AdaBN~\cite{schneider2020improving} & $50.55 \pm 0.01$ & $48.02 \pm 0.20$ & $39.65 \pm 0.10$ \\
T3A~\cite{iwasawa2021test} & $39.05 \pm 0.01$ & $39.28 \pm 0.03$ & $37.46 \pm 0.09$ \\
Tent~\cite{Wang2021} & $56.91 \pm 0.02$ & $54.66 \pm 0.20$ & $43.55 \pm 0.30$ \\
BACS~\cite{zhou2021bayesian} & $57.25 \pm 0.06$ & $55.40 \pm 0.27$ & $37.71 \pm 0.89$ \\
FR~\cite{eastwood2022sourcefree} & $51.22 \pm 0.02$ & $48.72 \pm 0.20$ & $40.11 \pm 0.08$ \\
Infomax~\cite{NIPS1991_a8abb4bb} & $\underline{58.22 \pm 0.03}$ & $\underline{56.40 \pm 0.19}$ & $\underline{45.80 \pm 0.29}$  \\ \hline
CAFe (w/o infomax) & $55.95 \pm 0.04$ & $53.13 \pm 0.14$ & $43.23 \pm 0.20$ \\
CAFe (dimwise) & $58.21 \pm 0.02$ & $56.35 \pm 0.20$ & $45.61 \pm 0.28$  \\
CAFe & ${\bf 58.76 \pm 0.04}$ & ${\bf 56.90 \pm 0.12}$ & ${\bf 46.92 \pm 0.34}$ \\ \hline
\end{tabular}
\caption{Test accuracy [\%] on ImageNet-C with mini-batched online TTA.}
\label{atab:online_imagenet_accuracy}
\end{table*}
In the experiment described in \secref{ssec:exp_setting}, we ran TTA on the target dataset for one epoch and then evaluated it on the same target dataset, where each sample in the target dataset had to be accessed twice.
However, there may be a situation in which we cannot access the whole target dataset at once.
In this section, we describe performing TTA in a mini-batched online manner.
Here, each target mini-batch can be accessed only one time.
In other words, for each mini-batch, we perform one step of TTA optimization and then make a prediction on the mini-batch.
This mini-batched online procedure is computationally efficient because we can discard the mini-batch after the prediction.

\tabref{atab:online_cifar_accuracy} and \tabref{atab:online_imagenet_accuracy} show the test accuracy on CIFAR10/100-C and ImageNet-C evaluated in a mini-batched online manner.
Compared with the offline TTA (\tabref{tab:accuracy}), the accuracies deteriorated in most cases because the target mini-batches were predicted with an insufficiently adapted model in the beginning.
However, CAFe had the best accuracy on the mixed set of CIFAR10/100-C and all sets of ImageNet-C and competitive accuracy in other cases.

\subsection{Adapting to Domain Shift}\label{assec:svhn}
\begin{table}[tb]
\centering
\begin{tabular}{l|l}\hline
Method & Accuracy [\%] \\ \hline
Source & $73.15$ \\
AdaBN~\cite{schneider2020improving} & $70.58 \pm 0.61$ \\
T3A~\cite{iwasawa2021test} & $76.31 \pm 0.10$ \\
Tent~\cite{Wang2021} & $78.03 \pm 0.36$ \\
BACS~\cite{zhou2021bayesian} & $86.10 \pm 0.29$ \\
FR~\cite{eastwood2022sourcefree} & $83.61 \pm 0.48$ \\
Infomax~\cite{NIPS1991_a8abb4bb} & $84.89 \pm 0.05$ \\ \hline
CAFe (w/o infomax) & $78.25 \pm 0.24$ \\
CAFe (dimwise) & ${\bf 87.65 \pm 0.35}$ \\
CAFe & $\underline{87.15 \pm 0.21}$ \\ \hline
\end{tabular}
\caption{Test accuracy on adapting SVHN to MNIST.}
\label{atab:svhn_accuracy}
\end{table}

We adapted the model pre-trained on SVHN~\cite{svhn} to MNIST~\cite{mnist}, which are often used in domain adaptation studies.
The details of these datasets are described in \asecref{asec:dataset_details}.
\tabref{atab:svhn_accuracy} shows that CAFe outperformed the other TTA methods and that CAFe (dimwise) performed the best.

\subsection{Adapting to the Same Distribution}\label{assec:adapting_same_distribution}
\begin{table*}[tb]
\centering
\begin{tabular}{l|lll}\hline
Method & CIFAR10 & CIFAR100 & ImageNet \\ \hline
Source & $90.52$ & $63.61 $ & $76.15$ \\ 
AdaBN~\cite{schneider2020improving} & $89.08 \pm 0.19$ & $61.12 \pm 0.32$ & $75.64 \pm 0.08$ \\ 
T3A~\cite{iwasawa2021test} & $90.43 \pm 0.04$ & $62.59 \pm 0.12$ & $71.70 \pm 0.05$ \\ 
Tent~\cite{Wang2021} & $89.41 \pm 0.05$ & $61.88 \pm 0.11$ & $75.97 \pm 0.10$ \\ 
BACS~\cite{zhou2021bayesian} & $89.58 \pm 0.04$ & $62.47 \pm 0.13$ & $76.38 \pm 0.05$ \\ 
FR~\cite{eastwood2022sourcefree} & $89.42 \pm 0.07$ & $61.59 \pm 0.13$ & $75.70 \pm 0.03$ \\
Infomax~\cite{NIPS1991_a8abb4bb} & $89.55 \pm 0.04$ & $62.28 \pm 0.07$ & $76.02 \pm 0.01$ \\ \hline
CAFe (w/o infomax) & $89.43 \pm 0.02$ & $61.66 \pm 0.08$ & $75.50 \pm 0.03$ \\ 
CAFe (dimwise) & $89.56 \pm 0.03$ & $62.28 \pm 0.06$ & $76.04 \pm 0.01$ \\ 
CAFe & $89.59 \pm 0.06$ & $62.30 \pm 0.13$ & $75.97 \pm 0.05$ \\ \hline
\end{tabular}
\caption{Test accuracy [\%] on the same distribution.}
\label{atab:in-distribution_accuracy}
\end{table*}

There may also be a situation in which the source and target distributions are the same, where we do not know in advance.
Here, we performed TTA on a target domain identical to the source one.
We adapted the source model pre-trained on CIFAR10/100 and ImageNet to their test split.
\tabref{atab:in-distribution_accuracy} shows that CAFe does not largely deteriorate accuracy as the baselines.

\subsection{Comparison to Unsupervised Domain Adaptation}\label{assec:domain_adaptation}
\begin{table}[tb]
\centering
{\small
\begin{tabular}{l|lll} \hline
Method & \shortstack{CIFAR10-C \\ (Mixed)} & \shortstack{CIFAR100-C \\ (Mixed)} & \shortstack{ImageNet-C \\ (All-mixed)}  \\ \hline
Source & $63.46\pm 0.61$ & $34.16\pm 0.20$ & $39.16\pm 0.01$ \\
DANN & $69.95 \pm 0.61$ & $38.63\pm 0.36$ & $45.42\pm 0.63$ \\
CAFe & ${\bf 70.06\pm 0.25}$ & ${\bf 40.01\pm 0.36}$ & ${\bf 48.55\pm 0.26}$  \\ \hline
\end{tabular}
}
\caption{Test accuracy [\%] on the target datasets adapted with DANN~\cite{ganin2016domain}.
The accuracies of the source model and CAFe are also shown for comparison.}
\label{atab:dann_accuracy}
\end{table}

We compared CAFe with unsupervised domain adaptation (UDA), which allows access to the source and target datasets simultaneously.
We used DANN~\cite{ganin2016domain}, a representative UDA method.
DANN employs a discriminator that takes the intermediate feature representations of the classifier and discriminates which domains they are from.
By training the classifier and the discriminator adversarially, the classifier is expected to be able to learn domain-invariant feature representations.
The discriminator we used is composed of three BN and fully-connected layers with ReLU activation, whose numbers of the hidden dimensions are 1,024.
We fine-tuned the source-pretrained classifiers with DANN.
For the CIFAR experiments, we set $\text{batch\_size}=128, \text{learning\_rate}=0.001$ and $\text{momentum}=0.8$.
For the ImageNet experiments, we set $\text{batch\_size}=64, \text{learning\_rate}=0.001$ and $\text{momentum}=0.8$.
We trained DANN for 400 iterations for CIFAR10/100-C and 20,000 iterations for ImageNet-C.
\tabref{atab:dann_accuracy} reports the best accuracy.
Compared with the TTA results in \tabref{tab:accuracy}, it is clear that DANN improved the target accuracy more than the source model did.
However, CAFe outperformed DANN.
This is because the discriminator of DANN has to learn from scratch.
The maximum number of iterations we trained for is one epoch based on the number of the source data.
One epoch is insufficient for the discriminator to learn the domain discrimination.
This suggests that DANN must be trained for a number of epochs for it to gain accuracy.

\end{document}